\documentclass[3p,times,authoryear, preprint]{elsarticle}

\usepackage[figuresright]{rotating}
\usepackage{graphicx} 
\usepackage{multirow}
\usepackage{subfiles}
\usepackage{multirow}
\usepackage{graphicx,tikz}
\usetikzlibrary{decorations.pathreplacing}
\usepackage{amsmath,amssymb}
\usepackage{xcolor}
\usepackage[colorlinks=true]{hyperref}

\begin{document}

\begin{frontmatter}


\title{FWin transformer for dengue prediction under climate and ocean influence}
\author[1]{Nhat Thanh Tran\corref{cor1}}
\ead{nhattt@uci.edu}
\author[1]{Jack Xin}
\author[3]{ Guofa Zhou}

\cortext[cor1]{Corresponding author}
\affiliation[1]{organization={Department of Mathematics, University of California, Irvine},    state={California}, country={USA}}
\affiliation[3]{organization={Department of Population Health and Disease Prevention, University of California, Irvine}, state={California}, country={USA}}





\address{}

\begin{abstract}
Dengue fever is one of the most deadly mosquito-born tropical infectious diseases. Detailed long range forecast model is vital in controlling the spread of disease and making mitigation efforts. In this study, we examine methods used to forecast dengue cases for long range predictions. The dataset consists of local climate/weather in addition to global climate indicators of Singapore from 2000 to 2019. We utilize newly developed deep neural networks to learn the intricate relationship between the features. The baseline models in this study are in the class of recent transformers for long sequence forecasting tasks. We found that a Fourier mixed window attention (FWin) based transformer performed the best in terms of both the mean square error and the maximum absolute error on the long range dengue forecast up to 60 weeks. 
\end{abstract}


\end{frontmatter}

\section{Introduction}

The modeling and forecasting of vector-borne diseases (VBD) 
 are scientifically challenging and important to the society at large due to their complex non-local temporal dependencies in the data as well
as external climate factors. VBD originated from tropical 
countries pose serious public health risks to both local and global communities. 
Among them, malaria and dengue fever are the two most deadly mosquito-borne tropical infectious diseases, with about 240 million malaria cases globally and 440,000 malaria deaths annually, and 50-100 million dengue cases. Currently, no tested vaccine or treatment is available to stop or prevent dengue fever. Thus modeling dengue disease evolution is particularly significant.
\medskip

The correlation of weather and climate with VBD evolution is well-documented  (\citep{zhou2004association,climate_dengue_13,Taiwan_dengue_2017,Central_Vietnam_dengue_2020,climate_vector_srilanka_2022} among others). 
With global warming upon us, 
higher temperatures create more habitats for mosquitoes
to infect unexposed human populations and 
spread diseases.
Just in July 2023, 
about 80 million Americans experienced a heat index of at least 
105 degrees according to the National Weather Service.
The extreme heat waves prompt irregular typhoons in Asia and 
flash floods in north America.
Such intense precipitation and flooding events become more frequent and longer (unless humans essentially stop adding carbon dioxide to the atmosphere), behaving as {\it strongly non-stationary} instead of traditional seasonal signals. They can favor mosquito breeding and survival to further complicate VBD  evolution.
Besides temperature, ocean currents can also influence the infection dynamics of VBD  \citep{Taiwan_dengue_2017,climate_vector_srilanka_2022}. 
\medskip

Knowledge of weather and population 
susceptibility cycles is known to help prediction, see \citep{Brazil_dengue_2021,LXZ_2022} and references therein. A new challenge of VBD modeling is to extract critical information from complex correlations and multiple factors {\it in the presence of intense transients lasting for week long periods due to extreme climate events}. Ensemble machine learning (ensemble support vector machines \citep{Brazil_dengue_2021}), recurrent neural networks (RNN) and regression \citep{LXZ_2022}, among other tools \citep{ML_Infectious_Disease_Review_2023} have been utilized in the past to study the effects of climate and seasonal weather. However due to either stationary hypothesis or simple decision boundary or one time unit ahead recursion, these existing tools are not well-suited for predicting the aftermath of intense non-stationary behavior in the data.
\medskip

In this paper, we study a {\it local-global attention} based {\it efficient transformer} \citep{fouriermixedtran2023} for {\it non-stationary VBD modeling and prediction in a specific spatial region}. For simplicity, we leave the spatial synchrony issue \citep{spatialpopsync99} (coupling among neighboring regions) to a future study.
Transformer networks equipped with attention blocks  \citep{vaswani2017attention} have powered the recent breakthroughs of natural language processing (NLP and Open AI's Chat-GPT) where long range temporal correlations exist. Since VBD data are not as abundant as in NLP, {\it light weight and efficient transformer networks} are more suitable and will be our main objects of study here. A successful strategy to reduce computational complexity of transformers is to approximate the full attention by a local attention (e.g. window attention \citep{Swin}) 
globalized by a subsequent mixing step (e.g. through shifting \citep{Swin} or shuffling \citep{shuffleT_2021} windows among other treatments). Recently, Fourier transform based mixing 
has been found competitive in accuracy and efficiency in both training and inference on long sequence time series forecasting tasks. On standard benchmarks \citep{informer_21} as well as highly non-stationary  power grid data \citep{glassoformer_22}, Fourier-mixed window attention (FWin, \citep{fouriermixedtran2023}) out-performs prob-sparse attention \citep{informer_21} and other recent models such as Autoformer \citep{autoformer_21}, Fedformer \citep{fedformer}, ETSformer \citep{etsformerwoo2022} and PatchTST \citep{patchtstnie2023}. We aim to continue this line of inquiry here on multi-variate dengue and climate data in Singapore which provide a real-world VBD data set to help understand and evaluate transformers as a new tool for advancing public health. 
\medskip

The innovations of this paper include a comprehensive transformer based generative model to encompass essential driving mechanisms of VBD evolution and make fast prediction of non-stationary dynamic behavior over a longer temporal duration than existing methods. The ideas of attention and multi-variate data fusion have been applied before in the context of infectious disease prediction, e.g. hand-foot-mouth disease and hepatitis
beta virus \citep{orient_attn_2022}, influenza and dengue \citep{attn_rnn_2019,vietnam_dengue_nguyen_et_al} etc. However, 1) the prior approaches of using a composition of a linear projection and softmax normalization as attention \citep{orient_attn_2022,attn_rnn_2019} or a special form of attention in machine translation \citep{luong2015effective} adopted in \citep{vietnam_dengue_nguyen_et_al} is not robust compared to the canonical quadratic (Q,K,V) form \citep{vaswani2017attention}; 2) these methods have not been evaluated on the public long sequence time-series forecast benchmarks \citep{informer_21}; and 3) the models by design can only make one time unit predictions which is not enough for any early warning. Though the attention enhanced LSTM in \citep{vietnam_dengue_nguyen_et_al} was extended from 1 month ahead to 2/3/6 month ahead predictions, the performance was increasingly worsening. Our method being a generative transformer can naturally make multi-time unit predictions.
\medskip

The rest of our paper is organized as follows. 
In section 2, we discuss the Singapore dengue data set and two prediction tasks. 
In section 3, we give an overview of transformer models in this paper including FWin transformer. In section 4, we compare results from the transformer models on the two prediction tasks and give our interpretations. 
In section 5, we provide an ablation study on the choice of window lengths of FWin transformer. 
Concluding remarks are in section 6.

\section{Dataset and Task}

\subsection{Data}
\begin{figure}[ht!]
    \centering
\includegraphics[width=\textwidth]{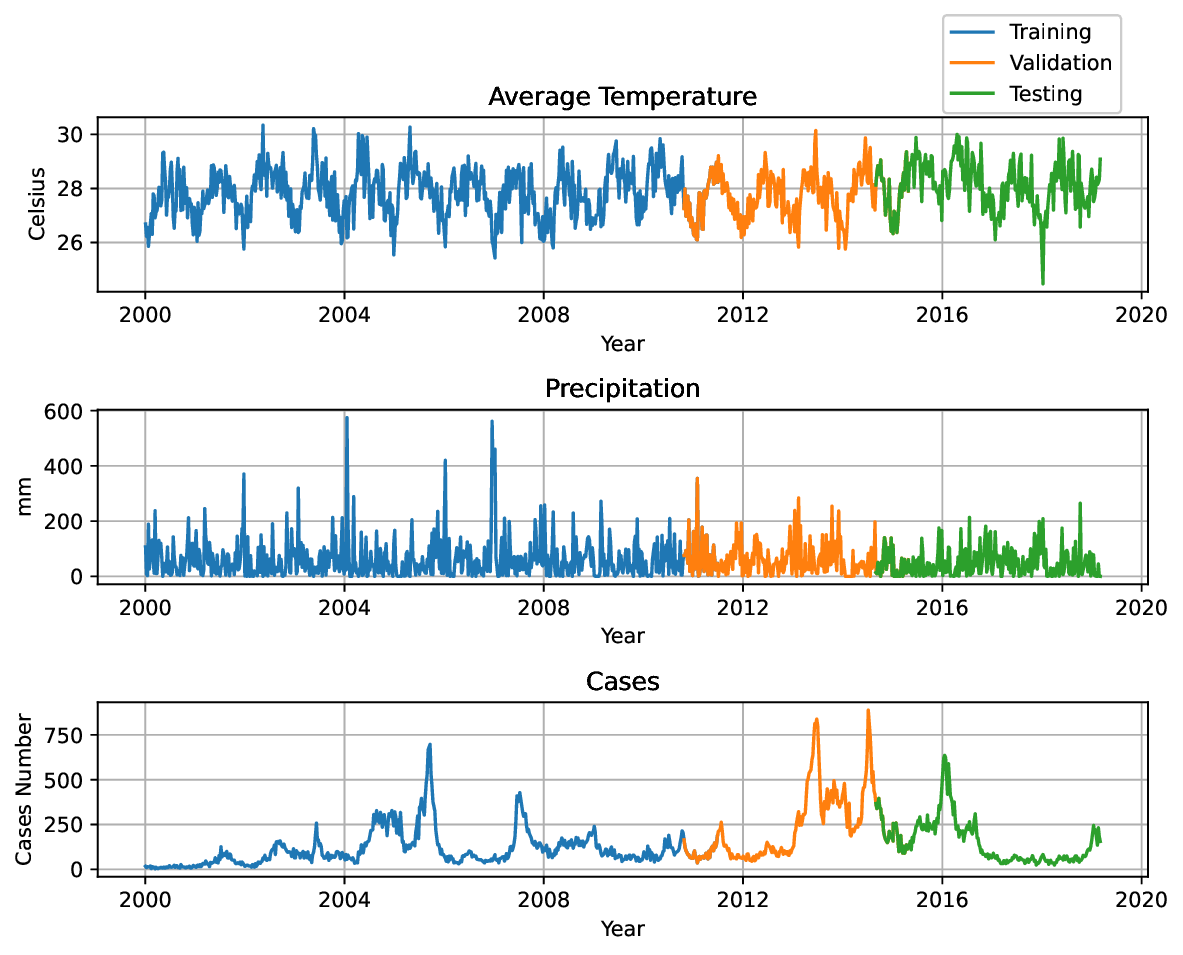}
    \caption{Sample features of the dataset. The plot includes the average temperature and precipitation from 2000 to 2019 with the dengue cases number. Here blue, orange, green indicate training, validation and testing split respectively.}
    \label{fig:Sample_weather_dataset}
\end{figure}

The dataset contains 1000 weeks of Singapore's weekly dengue data spanning from 2000 to 2019. The dataset's features include the following variables:  the cases number, average temperature, precipitation, Southern Oscillation Index (SOI), Oceanic Niño Index (ONI) total, ONI anomaly, Indian Ocean Dipole (IDO), IDO East, NINO1+2, NINO3, NINO4, NINO3.4, and the respective NINO anomaly. We refer to Tab. \ref{tab:data_feature_description} for a description of the features. Many of the features are reported in a daily or monthly interval. Whenever data are available at a coarser temporal resolution than weekly, we select the data corresponding to the month of the first day of that week. In cases where data is available at a finer temporal resolution than weekly, we calculate the average data for that week. The train/val/test split ratio is 6/2/2. We present a sample of the features of the dataset with the corresponding split ratio in Fig. \ref{fig:Sample_weather_dataset}. Data normalization applies to the entire dataset before passing it to the model. This means that each feature in the dataset will have zero mean and variance equal to 1. We label the data set as Singapore Dengue (SD). The processed data is available upon request.

\begin{table}[ht!]
    \caption{Summary of important features in the Singapore dataset.}
    \centering

\begin{tabular}{p{0.25\textwidth}|p{0.6\textwidth}}
     Feature& Description  \\
     \hline
     Average Temperature& Average temperature collected daily in Celsius. Weekly data is the total average temperature divide by the number of days. \\
     \hline
     Precipitation & Precipitation collected daily in mm. Total amount each week is combined during the pre-process. Missing data assumed to be zero. \\
     \hline
     Southern Oscillation Index & Compare the difference from average air pressure in the western Pacific, measured in Darwin, Australia, to the difference from average pressure in the central Pacific, measured at Tahiti \citep{soi_web}.\\
     \hline
     Oceanic Niño Index & Running 3-month average sea surface temperatures in the east-central tropical Pacific between 120W-170W \citep{oni_web}. \\
     \hline
     Indian Ocean Dipole &  Measure of the surface temperature difference between the western and eastern tropical Indian ocean \citep{iod_saji}. \\
     \hline
     NINO1+2 & Represent the sea surface temperature correponds with the region of coastal South America (0-10S, 90W-80W). \\
     \hline
     NINO3 & Represent the average sea surface temperature across the region (5N-5S, 150W-90W)\\
     \hline
     NINO3.4 & Represent the average equatorial sea surface temperature across the Pacific from about the dateline to the South American coast (5N-5S, 170W-120W).\\
     \hline
     NINO4 & Represent the sea surface temperature in the central equatorial Pacific (5N-5S, 160E-150W). \\
     \hline
\end{tabular}

    \label{tab:data_feature_description}
\end{table}


\subsection{Prediction Task}
\begin{figure}
    \centering
\begin{tikzpicture}[scale=0.75,every node/.style={transform shape}, thin]

  \def\numRows{3}
  \def\numCols{4}
  \def\cellSize{1}
  \def\cellSpace{0.2}

      
  \pgfmathtruncatemacro{\numRows}{3}
  \pgfmathtruncatemacro{\numCols}{4}

  \def\cellSize{0.25}




  \filldraw[orange!20, draw=white] (0,0) rectangle (4, -0.25);
  \filldraw[blue!20, draw=white] (0,-0.25) rectangle (4, -0.75);

  \draw[step=\cellSize, black!50] (0,0) grid (6, -0.75);


  \draw[step=\cellSize, black!50] (4,0) grid (6, -0.75);

    \draw[decorate,decoration={brace,amplitude=10pt},black] (0,0) -- (4, 0) node[midway, above=15pt,black] {$m$};
    
    \draw[decorate,decoration={brace,amplitude=10pt},black] (4,0) -- (6, 0) node[midway, above=15pt,black] {$n$};

    \node[] at (-1,-0.25) {Input};
    \node[] (a_input_node) at (3,-0.75){};

    \draw[] (2,-2) rectangle node[text=black] {Model} (4,-3);
    \node[] (a_model) at (3,-2) {};
    \node[] (a_model_below) at (3,-3) {};
    \draw[->] (a_input_node)--(a_model);

    \node[] (a_output) at (3,-4) {};
  \filldraw[orange!20, draw=white] (2,-4) rectangle (4, -4.25);

  \draw[step=\cellSize, black!50] (2-0.001,-4) grid (4, -4.25);
 \node[] at (-1,-4.1) {Output};

 \draw[->] (a_model_below) -- (a_output);
 \node[below] at (3, -5) {(a)};
 
  \filldraw[orange!20, draw=white] (10,0) rectangle (14, -0.25);
  \filldraw[blue!20, draw=white] (10,-0.25) rectangle (16, -0.75);
  
  \draw[step=\cellSize, black!50] (10-0.001,0) grid (16, -0.75);


    \draw[decorate,decoration={brace,amplitude=10pt},black] (10,0) -- (14, 0) node[midway, above=15pt,black] {$m$};
    
    \draw[decorate,decoration={brace,amplitude=10pt},black] (14,0) -- (16, 0) node[midway, above=15pt,black] {$n$};

    \node[] at (9,-0.25) {Input};
    \node[] (a_input_node) at (13,-0.75){};

    \draw[] (12,-2) rectangle node[text=black] {Model} (14,-3);
    \node[] (a_model) at (13,-2) {};
    \node[] (a_model_below) at (13,-3) {};
    \draw[->] (a_input_node)--(a_model);

    \node[] (a_output) at (13,-4) {};
  \filldraw[orange!20, draw=white] (12,-4) rectangle (14, -4.25);

  \draw[step=\cellSize, black!50] (12-0.001,-4) grid (14, -4.25);
 \node[] at (9,-4.1) {Output};

 \draw[->] (a_model_below) -- (a_output);

 \node[below] at (13, -5) {(b)};


\end{tikzpicture}

    \caption{(a) Input-Output structure of MS task. (b) Input-Output structure of MM task. Here orange shaded cells indicate non-zero value of the response feature, blue shaded cells indicate non-zero value of the predictor features, and white cells indicate zero padded value.}
    \label{fig:prediction_task}
\end{figure}
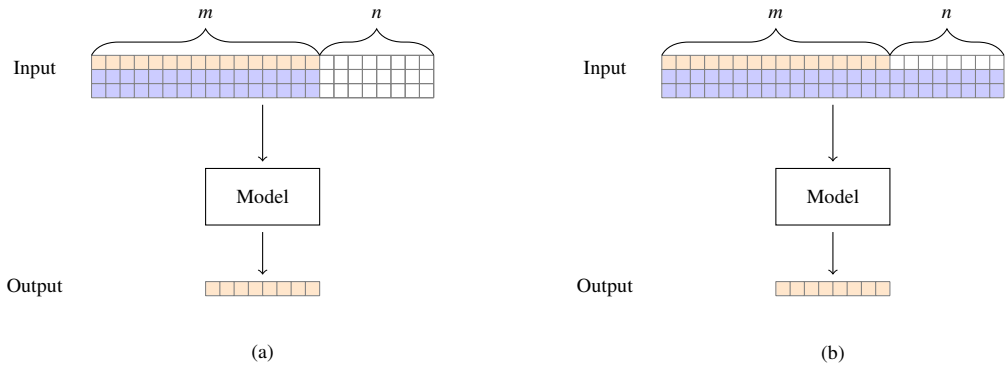

In this paper, we are interested in predicting the dengue cases number using all provided features. The appropriate task for this is Multivariate to Univariate (MS). For this prediction task, we utilize information from all features from the past $m$ time steps, and predict the number of dengue cases for the next $n$ time steps. In our model, we set $m$ to be 36 by default, and $n \in \{24, 36, 48, 60\}$. 
\medskip
Since we do not need to predict the other features, e.g. precipitation, we introduce a new prediction task where the inputs consists of $m+n$ times step of the predictor variables. However, the input will only contain $m$ time step of the response variable (RV), with the RV at the remaining $n$ time steps set to 0. We refer to this task as Modified Multivariate to Univariate (MM). The rationale behind this task is that if we have accurate forecasts for the predictor variables, then we can leverage this information to improve the prediction of the RV. Later, we will show that this modification increases the prediction power of our models. Figure \ref{fig:prediction_task} provides an overview of the overall structure of these tasks.

\section{Models}

Time scale latency between the effect of weather features (such as the abundance of water for larvae, and symptoms of disease in the host)  varies depending on the location. 
This latency could range from up to 6 months of delay to as short as just 6 weeks \citep{Titus_Bangladesh_data_mining}. Thus choosing a lag order (an integer parameter) in a standard statistical model such as ARIMA or VAR is non-trivial for our dataset. Moreover, traditional methods require choosing significant features, i.e. Pearson's correlation, before passing them in to the model \citep{idsignificantfeature_felistin}. Also features are assumed to be linearly correlated, which may not be true in general. In cases where a nonlinear model is necessary, we must have some functional form to describe the relationship between the climate features and cases number. To address this issue, we will utilize recent deep neural networks, in particular attention based neural networks. We will discuss in detail the structure in the following sections. Below are some reasons why a neural network may resolve some of the issues we have with traditional methods. First, a deep neural network has the ability to extract important features through learning. Second, we treat each time step of the input as a token passing into the neural network. Given a certain time step information (token), the attention mechanism allows particular token to focus on other tokens, this alleviates the need to pre-define a lag order. Of course, if one has a prior knowledge of a lag order, then it can be incorporated into the model such that one only allows tokens in that particular lag window to affect the final prediction. We will show this as an advantage of FWin. Lastly, other than choosing an architecture for a deep neural network, since in most cases it is a universal approximator, we do not need to have an exact equation to describe the nonlinear relationship between predictor and response variables.      
\medskip

In order to accomplish the dengue forecasting task, we will utilize some of the recently developed deep neural network models. We will compare the following models: FWin \citep{fouriermixedtran2023}, Informer 
 \citep{informer_21}, FEDformer \citep{fedformer}, Autoformer 
 \citep{autoformer_21}, ETSformer \citep{etsformerwoo2022}, PatchTST 
 \citep{patchtstnie2023}. 

\subsection{Background}
Many of the models presented in this paper utilize similar ideas from Transformer \citep{vaswani2017attention}. This was originally design for natural language tasks. In recent years, it was demonstrated that similar approach can be used for time series. We will provide some background information on these models next.

\subsubsection{Transformer}
Transformer is a basis for many of the newly developed models. Transformer has an encoder-decoder structure. The encoder maps an input $x$ to a representation $y$. Then the decoder accepts $y$ as an input to generate an output $z$. Encoder composes of stacks of self attention and feed forward layers. Similarly, decoder composes of stacks of masked self attention, cross attention, and feed forward layers \citep{vaswani2017attention}. Attention is an essential component of the Transformer, thus we will discuss them in more detail in the next section. 

\subsubsection{Attention}
Let $x\in \mathbb{R}^{L\times d}$ be the input sequence, where $L$ is the sequence length and $d$ is the feature dimension. Here $d$ loosely can be understood as the number of features in the dataset, usually the input is passed through a linear projection to project the original number of features to a larger dimension. We compute queries (Q), keys (K), values (V):
\begin{equation}
     Q = xW_Q + b_Q,\,
    K = xW_K + b_K,\,
    V = xW_V + b_V,
\end{equation}
where $W_Q, W_K, W_V\in \mathbb{R}^{d\times d}$ are the weighted matrix, and $b_Q, b_K, b_V\in\mathbb{R}^{L\times d}$ are the bias matrix. Attention is defined as:
\begin{equation}\label{eqn:attn_eqn}
     A(Q, K, V) = \text{softmax}(QK^T/\sqrt{d}) V,
\end{equation}
where $A$ is the attention function and softmax function apply along the row dimension \citep{vaswani2017attention}. We usually refer to this calculation as full self attention, because the queries, keys, and values are linear projections of the same input, and there is a full matrix multiplication of $Q$ and $K^T$. However, for cross attention, the query comes from the linear projection of decoder input, while the keys and values come from encoder output. For causality, masked attention is a restriction to the self attention calculation where we prohibit the interaction of the current query to future keys.
\medskip

For window attention \citep{Swin, fouriermixedtran2023}, one computes the attention on sub-sequences then concatenate the results together. We first divide sequence $x$ into $N$ subsequences: $x^{(1)}, x^{(2)},\dots, x^{(N)}$, such that $x = [x^{(1)}, x^{(2)},\dots, x^{(N)}]^T$. Each $x^{(i)}\in \mathbb{R}^{L/N\times d}$ for $i=1,2,\dots, N$, where $N=L/W$, $W$ is a fixed window size. This implies we divide the queries, keys and values as follow $Q = [Q^{(1)}, Q^{(2)},\dots, Q^{(N)}]^T$, $K = [K^{(1)}, K^{(2)},\dots, K^{(N)}]^T$, $V = [V^{(1)}, V^{(2)},\dots, V^{(N)}]^T$. Thus we compute attention for each subsequence as follows: 
\begin{equation}
    A(Q^{(i)},K^{(i)},V^{(i)}) = \text{softmax}\left(\,Q^{(i)}K^{(i)T}/\sqrt{d}\, \right) V^{(i)}.
\end{equation}
After computing the attention for each sub-sequence, we concatenate the sub-attentions to form the window attention:

\begin{equation}
A_w(Q,K,V) = 
\begin{bmatrix}
    A(Q^{(1)},K^{(1)},V^{(1)}) \\
    \vdots\\ 
    A(Q^{(N)},K^{(N)},V^{(N)})
\end{bmatrix}.    
\end{equation}

\subsection{Fourier Mix}
The FWin model uses window attention in place of full attention to reduce the computational complexity. However, this limits the interactions between the tokens, which may lead to degradation in performance. To resolve this issue, FWin utilizes the fast Fourier transform as a way to mix tokens among the windows. Given input $x\in \mathbb{R}^{L\times d}$, one computes Fourier transform along the feature dimension ($d$),  then along the time dimension ($L$), finally taking real part to arrive at:
\begin{equation}
    y = \mathcal{R}(\mathcal{F}_{\text{time}}(\mathcal{F}_{\text{feature}}(x))),
\end{equation}
where $\mathcal{F}$ is 1D discrete Fourier transform (DFT), and $\mathcal{R}$ is the real part of a complex number \citep{fouriermixedtran2023}.

\subsection{FWin Overview}
\begin{figure*}[ht]
    \centering

\begin{tikzpicture}[scale=0.75,every node/.style={transform shape}, thin]

\node[rectangle, draw,minimum width=2cm, minimum height=1cm, fill=blue!20] at (10.5,-1.5) (encin){Encoder Input};

\node[rectangle, draw,minimum width=2cm, minimum height=1.5cm, fill=blue!20] at (10.5,1) (encattn){\begin{tabular}{c}
     Multi-head  \\
     Window\\
     Self-Attention
\end{tabular}};

\node[rectangle, draw,minimum width=2cm, minimum height=1.5cm, fill=blue!20] at (10.5,3) (encdistil){\begin{tabular}{c}
     Distilling  \\
     Operation
\end{tabular}};

\node[rectangle, draw,minimum width=2cm, minimum height=1.5cm, fill=blue!20] at (10.5,5) (encfnet){\begin{tabular}{c}
     Fourier Mix
\end{tabular}};

\node[minimum width=2cm, minimum height=1.5cm] at (10.5,6) (encoder){\begin{tabular}{c}
     \textbf{Encoder}
\end{tabular}};

\node[rectangle, draw,minimum width=3cm, minimum height=7cm] at (10.5,3) (encblock){};

\node[rectangle, draw,minimum width=2cm, minimum height=1.5cm, fill=blue!20] at (10.5,7.5) (encfeat){\begin{tabular}{c}Concatenated \\Feature Map \end{tabular}};


\node[rectangle, draw,minimum width=2cm, minimum height=1cm, fill=blue!20] at (14.5,-1.5) (decin){Decoder Input};

\node[rectangle, draw,minimum width=2cm, minimum height=1.5cm, fill=blue!20] at (14.5,1) (decsattn){\begin{tabular}{c}
     Masked \\Multi-head  \\
     Window\\
     Self-Attention
\end{tabular}};

\node[rectangle, draw,minimum width=2cm, minimum height=1.5cm, fill=blue!20] at (14.5,3) (decfmix){\begin{tabular}{c}
     Fourier Mix
\end{tabular}};

\node[rectangle, draw,minimum width=2cm, minimum height=1.5cm, fill=blue!20] at (14.5,5) (deccattn){\begin{tabular}{c}
     Multi-head  \\
     Window\\
     Cross Attention
\end{tabular}};

\node[minimum width=2cm, minimum height=1.5cm] at (14.5,6) (encoder){\begin{tabular}{c}
     \textbf{Decoder}
\end{tabular}};

\node[rectangle, draw,minimum width=3.4cm, minimum height=7cm] at (14.5,3) (decblock){};

\node[rectangle, draw,minimum width=2cm, minimum height=1cm, fill=blue!20] at (14.5,7.5) (decfcl){Fully Connected Layer};

\node[rectangle, draw,minimum width=2cm, minimum height=1cm, fill=blue!20] at (14.5,9.0) (decout){Output};

\draw[-stealth] (encin) -- (encblock);
\draw[-stealth,scale=0.5] (encattn) -- (encdistil);
\draw[-stealth,scale=0.5] (encdistil) -- (encfnet);
\draw[-] (encblock) -- (encfeat);
\draw[-stealth,scale=0.5] (decin) -- (decblock);
\draw[-stealth,scale=0.5] (decsattn) -- (decfmix);
\draw[-stealth,scale=0.5] (decfmix) -- (deccattn);
\draw[-stealth,scale=0.5] (decblock) -- (decfcl);
\draw[-stealth,scale=0.5] (decfcl) -- (decout);

\draw[-stealth] (encfeat) --(12.5,7.5)--(12.5,5)-- (deccattn);


\end{tikzpicture}
\caption{FWin model overview \citep{fouriermixedtran2023}.}
\label{fig:model_overview}
\end{figure*}
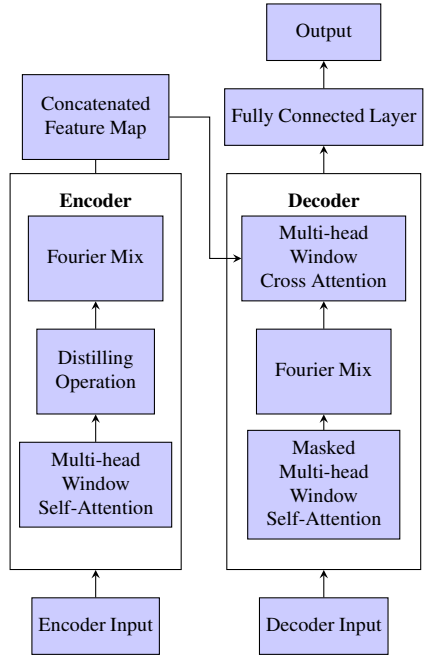

FWin performs the best among all of the models; and in addition, the procedure is general across different models. We will present FWin model structure in more detail. See Fig. \ref{fig:model_overview} for the model overview.
FWin is an adaptation of Informer \citep{informer_21} by employing the window attention \citep{Swin} mechanism to capture the local information and Fourier Mix layer \citep{Fnet} to mix tokens among the windows.  FWin has an encoder-decoder structure. First, raw input passes into the Encoder Input layer to embed the time, and positional information. In the encoder, the input first passes through the window attention, then dimensional reduction layer of Distilling Operation. This layer composes of convolution and MaxPool operations. Lastly, the tokens mix by the Fourier Mix layer, and go toward to the decoder. Second, the raw input passes into the Decoder Input layer with the time and positional information added into the input. In the decoder, the input passes through a masked window attention to respect causality. Then the tokens mix by the Fourier Mix layer, then pass  through the cross attention block,
and finally through a Fully Connected Layer (i.e. a linear projection to return an output with correct dimension) to produce the output. The input and output dimensions are the same as shown in Fig. \ref{fig:model_overview}.
\medskip

FWin employs window attention instead of full attention. One interpretation of this is that we limit token interactions in a small window. This implies that selecting a window size has a similar effect to choosing a lag order in a vector autoregression (VAR) model. We will delve into this effect in more detail in the ablation study section.

\subsection{Other Models}
FWin is derived from the Informer whose structure is similar to Fig. \ref{fig:model_overview}. The main difference is that Informer uses the so called Probsparse attention instead of Window attention and Fourier Mix. Probsparse attention \citep{informer_21} relies on a sparse query measurement function (an analytical approximation of Kullback-Leibler divergence) so that each key attends to only a few top queries.
\medskip

FEDformer uses an encoder-decoder structure as well, however, instead of canonical attention, it uses the frequency enhanced attention. It applies Fourier transform to the input, then select a few modes to compute attention. In addition, it incorporates a seasonal-trend decomposition layers to capture the global properties of time series \citep{fedformer}. Similarly, Autoformer uses Series Decomposition and Auto-Correlation instead \citep{autoformer_21}.  
\medskip

ETSformer uses similar structure with frequency attention and exponential smoothing attention to extract growth and seasonal information from the inputs. Here the attention is weighted with an exponential term that favors nearby tokens. Then it uses such information selection in the decoder to forecast the future horizon \citep{etsformerwoo2022}. 
\medskip

Lastly, PatchTST only uses the encoder structure of the Transformer. It first divides the input into patches, and passes each feature independently through the Transformer's encoder, then concatenate the outputs to form the final prediction \citep{patchtstnie2023}. 
\medskip

Some of the models only make multivariate prediction. Therefore, to generate univariate predictions, we simply extract the response feature from the model's output. This approach is standard for these types of problems. 

\subsection{Models Hyperameters}
For all of the models in this paper, we used their default hyper-parameters. We present the hyper-parameters used in FWin and Informer in Tab. \ref{tab:Model_hyperparameter}. In our experiments, we averaged the results over 5 runs. The total number of epochs is 6 with early stopping. We used the Adam optimizer, and the learning rate starts at $1e^{-4}$, decaying two times smaller every epoch. For ETSformer we used the initial learning rate of $1e^{-3}$ in the  exponential with a warm up learning rate schedule. For PatchTST we used the constant learning rate of $2.5e^{-3}$ and 100 training epochs with early stopping.   
\begin{table}[ht]
    \caption{Model default parameters for FWin and Informer \citep{fouriermixedtran2023}.}
    \centering
    \begin{tabular}{c c|c c}
        $d$ & 512  & Window size & 12\\
        $d_{ff}$& 2048& Cross Attn Window no. & 3 \\
        n\_heads & 8 & Epoch & 6\\
        Dropout & 0.05& Early Stopping Counter & 3\\
        Batch Size& 32& Initial Lr & 1e-4\\
        Enc.Layer no.& 2 & Dec.Layer no.& 1\\
    \end{tabular}

    \label{tab:Model_hyperparameter}
\end{table}

\section{Results and Discussion}

We present a summary of all the prediction tasks on the Singapore dataset in Tab. \ref{tab:Model_accuracy_Singapore}.  MAE $=\frac{1}{n}\sum_{i=1}^n |y- \hat{y}|$  and MSE $=\frac{1}{n}\sum_{i=1}^n (y- \hat{y})^2$ serve as evaluation metrics. The best results are highlighted in boldface, and the total count at the bottom of the table indicates how many times a particular method outperforms the others per metric per task. From Tab. \ref{tab:Model_accuracy_Singapore}, we observe that FWin has the best performance on both tasks. 
\medskip

In addition, we also demonstrate that including future predictor feature information (e.g. climate and ocean current features) in the model increases FWin performance (on dengue cases) significantly. We observed that for longer time forecasting, i.e. metrics of 36, 48, 60, the error for MM task is lower than MS task. However, for Informer and PatchTST, the models' performances decrease with additional information. An explanation for the degradation in performance is as follows: 1) the Informer's probspare attention only chooses a number of queries to compute attention, which means if the input length increases then a few top queries may not capture the majority of the contribution. 2) For PatchTST, it initally patches the input with overlaps, thus increasing the sequence length may cause too much overlap, leading to an overflow of information. In addition, it treats each feature independently, thus adding additional information for the predictor features may not affect the response feature. In case of Autoformer and ETSformer, including future predictor (e.g. climate and ocean current) information is beneficial to model prediction power. In particular, ETSformer's error reduces significantly for MM task compared to the MS task. However, FEDformer was not able to perform the MM task because the requirement of the input length of the decoder need to be half of the input of the sequence. 
\medskip

\begin{figure}[ht!]
    \centering
    \includegraphics[width=\textwidth]{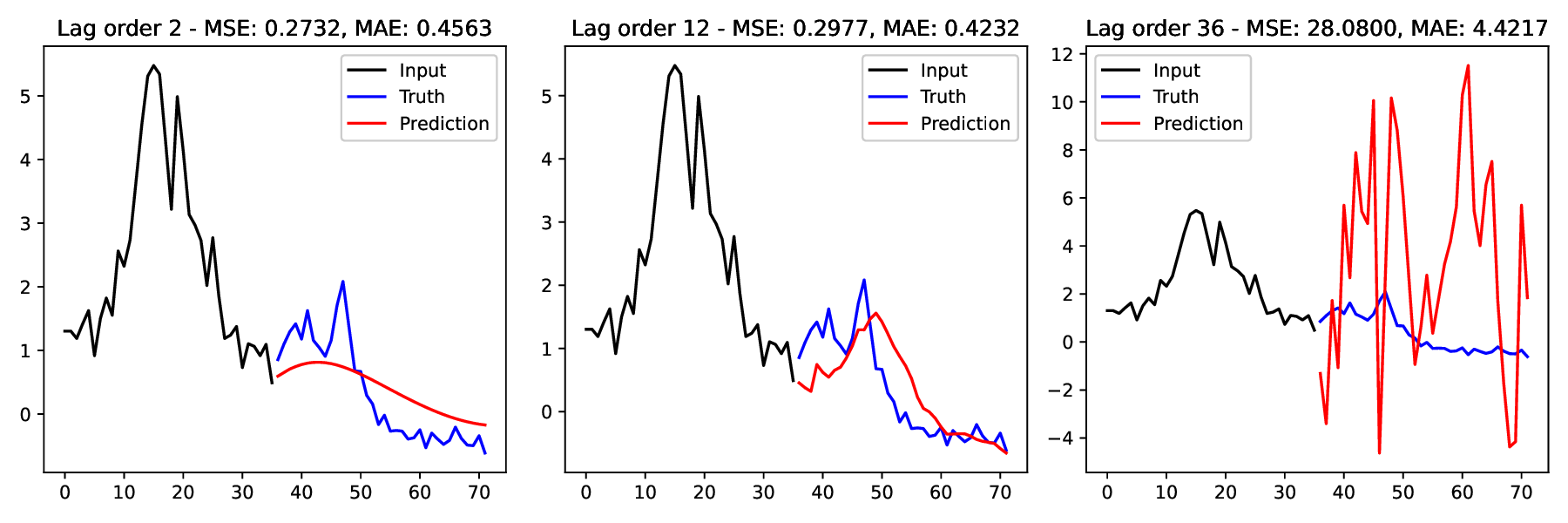}
    \caption{Sample VAR predictions with various lag orders. The $x$-axis represents time in weeks, and $y$-axis is the normalized cases number. The suptitle includes the lag order and the prediction errors. In black is the case number input, blue is the ground truth and red is the prediction of the model.}
    \label{fig:k58_var_predictions}
\end{figure}

Furthermore, we compared the vector autoregression (VAR) to  DNN models. VAR depends on a choice of lag order, which is non-trivial. Opting for a large lag order can lead to the model making wild predictions, resulting in large errors. On the other hand, selecting a small lag order can significantly reduce errors, but the prediction may appear too smooth, which is unrealistic given the behavior of the dataset. In Fig. \ref{fig:k58_var_predictions}, we provide a sample prediction of VAR. In this example, with a large lag order of 36 (the largest lag order in our setup), the prediction exhibits a large error. On the other extreme, with a lag order of 2, the error is significantly lower, however the prediction is smooth. For a medium lag order of 12, the prediction appears more realistic, however the error increases. Additionally, the overall error was higher than FWin, thus we do not present the full VAR results here. We called the VAR library from the package statsmodels for the simulations in this paper. 

\medskip

\begin{figure}[ht!]
    \centering
    \includegraphics[width=\textwidth]{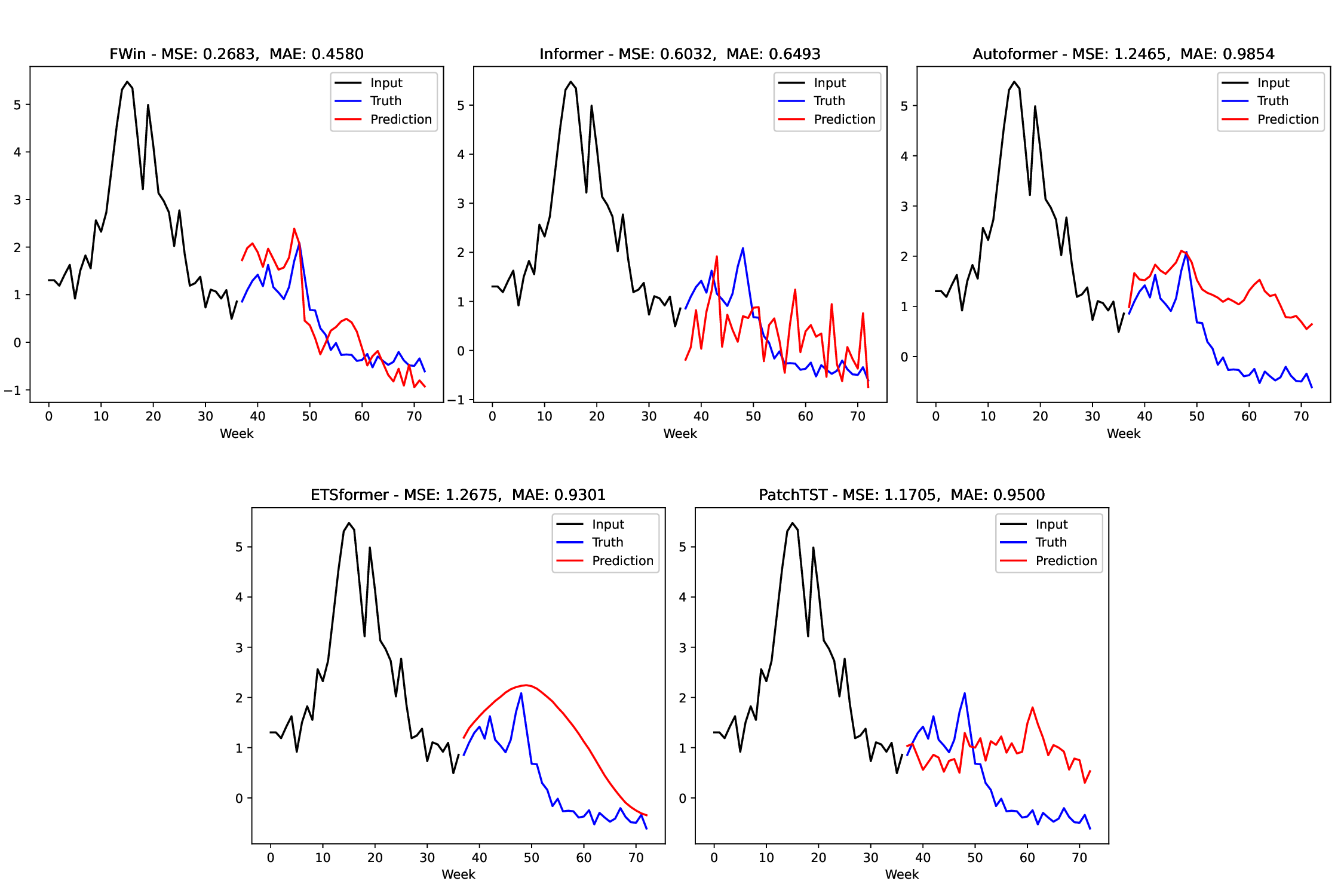}
    \caption{Sample models prediction for MM task. The $x$-axis represents time in weeks, and $y$-axis is the normalized cases number. The suptitle includes the model name and the prediction errors (MSE and MAE). In black is the case number input, blue is the ground truth and red is the prediction of the model.}
    \label{fig:k58_models_prediction_mms}
\end{figure}

We present a sample of the prediction of various models in Fig. \ref{fig:k58_models_prediction_mms} for the MM tasks. The $x-$axis represents the timescale in weeks, and $y-$axis is the normalized number of cases. FWin performs the best in terms of metrics presented and visual. It was able to predict a large drop in number of cases while many other models unable to do so. We noted that in the MM task, the model input is richer, containing the most information. This confirm the results we obtained in Tab. \ref{tab:Model_accuracy_Singapore}. 

\begin{table}[ht]
\caption{Accuracy comparison on Singapore data with input length of 36, best results highlighted in bold. Here MS, MM are multivariate to univariate, and modified multivariate to univariate respectively.``-" indicates the method is inapplicable for the task.}

    \centering
     \fontsize{8}{10}\selectfont
    \begin{tabular}{|c|c|cc|cc|cc|cc|cc|cc|cc|cc|cc|}
    \hline
         \multicolumn{2}{|c|}{Methods}& \multicolumn{2}{c|}{FWin} & \multicolumn{2}{c|}{Informer} & \multicolumn{2}{c|}{FEDformer} & \multicolumn{2}{c|}{Autoformer} & \multicolumn{2}{c|}{ETSformer} & \multicolumn{2}{c|}{PatchTST} \\   
         \hline
         \multicolumn{2}{|c|}{Metric} & MSE & MAE & MSE & MAE & MSE & MAE & MSE & MAE & MSE & MAE & MSE & MAE\\
         \hline
         \hline
         \multirow{4}{1em}{\rotatebox{90}{SD (MS)} } 
         & 24& \textbf{1.170}& \textbf{0.684}& 1.842& 0.877& 2.090& 1.085& 2.237& 1.127& 1.611& 0.856 & 1.273 & \textbf{0.684}\\
         & 36& \textbf{1.450}& \textbf{0.725}& 1.974& 0.942& 2.455& 1.187& 2.441& 1.198& 1.823& 0.938 & 1.664 & 0.815\\
         & 48& \textbf{1.782}& \textbf{0.830}& 2.043& 0.958& 2.912& 1.334& 2.927& 1.370& 2.271& 1.063 & 1.887 & 0.915\\
         & 60& \textbf{1.518}& \textbf{0.826}& 1.784& 0.910& 2.938& 1.358& 3.015& 1.395& 2.379& 1.128 & 2.390 & 1.107\\
         \hline
         \multirow{4}{1em}{\rotatebox{90}{SD (MM)} } 
         & 24& \textbf{1.303}& \textbf{0.787}& 2.137& 1.002& - & - & 1.951& 1.055& 1.480& 0.885 & 1.734 & 0.833\\
         & 36& \textbf{1.156}& \textbf{0.680}& 2.043& 0.968& - & - & 2.136& 1.105& 1.305& 0.774 & 1.994 & 0.937\\
         & 48& \textbf{1.251}& \textbf{0.693}& 2.228& 1.033& - & - & 2.238& 1.152& 1.355& 0.825 & 2.352 & 1.062\\
         & 60& \textbf{1.124}& \textbf{0.710}& 1.911& 0.946& - & - & 2.680& 1.272& 1.295& 0.781 & 2.579 & 1.117\\
         \hline
         \hline
         \multicolumn{2}{|c|}{Count}& \multicolumn{2}{c|}{16}& \multicolumn{2}{c|}{0}& \multicolumn{2}{c|}{0} &  \multicolumn{2}{c|}{0}& \multicolumn{2}{c|}{0}& \multicolumn{2}{c|}{1}   \\
         \hline
    \end{tabular}

\label{tab:Model_accuracy_Singapore}
\end{table}

\section{Ablation Study}

In this section, we will examine the effect of window size of the FWin model on its performance. We can think of the window size of FWin as prior knowledge of the time delay effect of weather/climate on the dengue cases. Due to the restricted interaction within each window, the cases number in a particular window only attends to local weather information. For this experiment, we utilize FWin in the MS task, with window sizes of 1, 2, 4, 6, 12, and 18. 
\medskip

From Tab. \ref{tab:windowsize_vs_metric}, we observe that the biggest window size of 18 gives the best performance in most cases. This is intuitively consistent with the design of FWin where we expect that the larger the window size, the better the overall performance. In addition, we observe that for the smallest window size of 1, the errors are significantly lower than naively expected. In particular, under the MAE metric, the window size of 18 performs the best, while under the MSE metric, a window size of 1 is the best. MSE amplifies the effects of large errors while suppressing those from the small errors. Thus under this metric, the smallest window size of 1 does not result in many large errors overall. On the other hand, MAE penalizes all error types equally, thus implying that the window size of 18 mostly results in small errors. We observe from the proof of FWin attention equivalency with the full attention (Theorem 5.6 in \citep{fouriermixedtran2023}) that the theorem is true without any assumption on the structure of the attention matrix if the window size is 1. This provides an explanation for why we encountered low MSE error for window size of 1, i.e. the model does not make many large errors in the prediction.

\begin{table}[ht]
\caption{Accuracy comparison of FWin model using different window sizes with input length of 36. Best results highlighted in bold.}
    \centering
     \fontsize{8}{10}\selectfont
    \begin{tabular}{|c|c|cc|cc|cc|cc|cc|cc|cc|cc|cc|}
    \hline
         \multicolumn{2}{|c|}{Window Size}&\multicolumn{2}{c|}{1}& \multicolumn{2}{c|}{2}& \multicolumn{2}{c|}{4} & \multicolumn{2}{c|}{6} & \multicolumn{2}{c|}{12} & \multicolumn{2}{c|}{18} \\   
         \hline
         \multicolumn{2}{|c|}{Metric} & MSE & MAE & MSE & MAE & MSE & MAE & MSE & MAE  & MSE & MAE & MSE & MAE\\
         \hline
         \hline
         \multirow{4}{1em}{\rotatebox{90}{SD (MS)} } 
         & 24& 1.195& 0.705& 1.282& 0.720& 1.284& 0.877& 1.223& 0.703& \textbf{1.181}& 0.678& 1.196& \textbf{0.667} \\
         & 36& \textbf{1.432}& 0.742& 1.507& 0.754& 1.517& 0.755& 1.492& 0.747& 1.458& 0.725& 1.440& \textbf{0.708} \\
         & 48& \textbf{1.673}& 0.828& 1.854& 0.866& 1.821& 0.859& 1.836& 0.862& 1.765& 0.824& 1.756& \textbf{0.817} \\
         & 60& \textbf{1.538}& 0.848& 1.552& \textbf{0.805}& 1.555& 0.810& 1.622& 0.830& 1.592& 0.813& 1.598& 0.809 \\
         \hline
         \hline
         \multicolumn{2}{|c|}{Count}& \multicolumn{2}{c|}{3}& \multicolumn{2}{c|}{1}& \multicolumn{2}{c|}{0}& \multicolumn{2}{c|}{0} &  \multicolumn{2}{c|}{1}&  \multicolumn{2}{c|}{3} \\
         \hline
    \end{tabular}

\label{tab:windowsize_vs_metric}
\end{table}
\medskip
In addition, we performed further experiments to understand the effect of shorter input sequence length to model performance. For this purpose, we reduced the input sequence length of the model from the default value of 36 to 18. Tab. \ref{tab:Model_accuracy_Singapore_SmallerInput} presents the results for all the transformer models under the MS task. We observe that for the shortest prediction length, PatchTST performs the best. However, for longer prediction lengths, FWin exhibits the best results. In general, the errors remain similar to their counterparts when the input length is 36. Therefore FWin is robust under the variation to shorter input lengths.

\begin{table}[ht]
\caption{Accuracy comparison on Singapore data with input length of 18, best results highlighted in bold. Here MS is multivariate to univariate task.}
    \centering
     \fontsize{8}{10}\selectfont
    \begin{tabular}{|c|c|cc|cc|cc|cc|cc|cc|cc|cc|cc|}
    \hline
         \multicolumn{2}{|c|}{Methods}& \multicolumn{2}{c|}{FWin} & \multicolumn{2}{c|}{Informer} & \multicolumn{2}{c|}{FEDformer} & \multicolumn{2}{c|}{Autoformer} & \multicolumn{2}{c|}{ETSformer} & \multicolumn{2}{c|}{PatchTST} \\   
         \hline
         \multicolumn{2}{|c|}{Metric} & MSE & MAE & MSE & MAE & MSE & MAE & MSE & MAE & MSE & MAE & MSE & MAE\\
         \hline
         \hline
         \multirow{4}{1em}{\rotatebox{90}{SD (MS)} } 
         & 24& 1.388& 0.673& 1.899& 0.888& 1.650& 0.890& 2.237& 1.127& 1.476& 0.839 & \textbf{1.040} & \textbf{0.621}\\
         & 36& \textbf{1.444}& \textbf{0.698}& 1.903& 0.921& 2.169& 1.021& 2.441& 1.198& 1.722& 0.908 & 1.542 & 0.772\\
         & 48& \textbf{1.564}& \textbf{0.771}& 2.045& 0.983& 2.863& 1.241& 2.927& 1.370& 2.147& 1.048 & 1.972 & 0.932\\
         & 60& \textbf{1.281}& \textbf{0.666}& 1.811& 0.928& 2.992& 1.282& 3.015& 1.395& 2.169& 1.073 & 2.282 & 1.050\\
         \hline
         \hline
         \multicolumn{2}{|c|}{Count}& \multicolumn{2}{c|}{6}& \multicolumn{2}{c|}{0}& \multicolumn{2}{c|}{0} &  \multicolumn{2}{c|}{0}& \multicolumn{2}{c|}{0}& \multicolumn{2}{c|}{2}   \\
         \hline
    \end{tabular}

\label{tab:Model_accuracy_Singapore_SmallerInput}
\end{table}

\section{Conclusions}

In this paper, we evaluated various attention-based deep neural networks for predicting dengue cases in Singapore. The dataset contains multiple features, including average temperature, precipitation, and global climate indices such as Southern Oscillation Index, Oceanic Ni$\Tilde{\text{n}}$o Index, and Indian Ocean Dipole. Among the models investigated in the study, FWin demonstrated the highest prediction accuracy overall. 
Moreover, incorporating future information in the modified multivariate to univariate task generally improved dengue case prediction for many models considered.
\medskip

For future work, we plan to incorporate a more explicit
climate and disease correlation layer (with help of certain prior knowledge) into the FWin model to enhance its performance. In addition, we plan to develop spatial-temporal transformer models that take into account geographical information and predict disease cases over multiple regions in countries bordering the Indian and Pacific oceans.


\section*{Authors contribution}

\textbf{Nhat Thanh Tran}: Methodology, Software, Validation, Investigation, Data Curation, Writing - Original Draft, Visualization.
\textbf{Jack Xin:} Conceptualization, Resources, Supervision, Writing - Review \& Editing, Funding acquisition. \textbf{Guofa Zhou:} Data Curation, Resources.

\section*{Acknowledgements}
The work was partially supported by 
NSF grants DMS-2219904 and DMS-2151235.

\newpage
\bibliography{bib}

\end{document}